\documentclass[11pt,twocolumn,letterpaper]{article}

\usepackage[left=0.8in,right=0.8in,top=0.8in,bottom=0.8in]{geometry}

\usepackage{times}
\usepackage{epsfig}
\usepackage{graphicx}
\usepackage{amsmath}
\usepackage{amssymb}

\usepackage{url}
\usepackage{svg}
\usepackage{multirow}
\usepackage{subcaption}
\usepackage{dblfloatfix}

\def\smean#1{\overline{#1}}
\def\svar#1{\overline{\overline{#1}}}

\newcommand{\ie}{\textit{i.e.,}}
\newcommand{\etal}{\textit{et al.}}

\begin{document}

\title{This Face Does Not Exist... But It Might Be Yours! \\ Identity Leakage in Generative Models\footnote{Pre-print of the paper to be presented at WACV 2021}}

\author{Patrick Tinsley \hspace{2cm} Adam Czajka \hspace{2cm} Patrick Flynn \\
University of Notre Dame\\
{\tt\small \{ptinsley,aczajka,flynn\}@nd.edu}
}

\date{}

\maketitle

\begin{abstract}
Generative adversarial networks (GANs) are able to generate high resolution photo-realistic images of objects that ``do not exist.'' These synthetic images are rather difficult to detect as fake. However, the manner in which these generative models are trained hints at a potential for information leakage from the supplied training data, especially in the context of synthetic faces. This paper presents experiments suggesting that identity information in face images can flow from the training corpus into synthetic samples without any adversarial actions when building or using the existing model. This raises privacy-related questions, but also stimulates discussions of (a) the face manifold's characteristics in the feature space and (b) how to create generative models that do not inadvertently reveal identity information of real subjects whose images were used for training. We used five different face matchers (face\_recognition, FaceNet, ArcFace, SphereFace and Neurotechnology MegaMatcher) and the StyleGAN2 synthesis model, and show that this identity leakage {\it does} exist for some, but not all methods. So, can we say that these synthetically generated faces truly do not exist? Databases of real and synthetically generated faces are made available with this paper to allow full replicability of the results discussed in this work.
\end{abstract}

\begin{figure}
    \centering
    \includegraphics[width=0.48\textwidth]{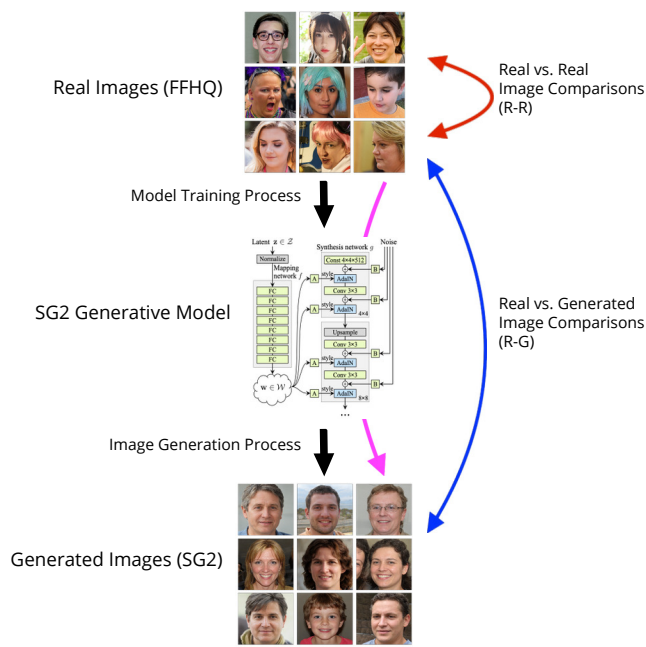}
    \caption{The proposed identity leakage path through the StyleGAN2 generator follows the pink line from real images (FFHQ) to generated images (SG2).}
    \label{fig:overview}
\end{figure}


\section{Introduction}

Since generative adversarial networks (GANs) were first introduced in 2014 \cite{goodfellow2014generative}, they have demonstrated impressive photo-realistic image-generating capabilities. During the GAN training process, a generator and a discriminator are pitted against each other for the purpose of elevating the generator. In the context of image data, these two adversaries battle over the production and recall of ``fake" images; the generator attempts to fool the discriminator by producing images that mimic the real data distribution while the discriminator aims not to be fooled by comparing and judging fake images from real. At the end of this minimax game, the resulting generator, having learned the general distribution of the supplied training data, can be used to synthesize never-before-seen ``fake'' images that adopt the characteristics of real face images while not appearing in the training data. The results and future promise of these generative models have led to heavy academic intrigue \cite{radford2015unsupervised, gulrajani2017improved}. However, there is a potential risk regarding their ability to generate images to establish fake identities that can be exploited. As a phenomenon closely related to the well-known ``fake news'' problem bedeviling rational discourse in politics, science, and society, the risks of identity synthesis are analogous to the risks of baseless information synthesis and deserve careful attention from the originating domain as well as the discipline of ethics \cite{shen2019fake}.

Of primary interest in this paper is an artifact of generative models that concerns individuals' sense of privacy regarding personal image data: {\bf identity leakage in generative models}. These models optimize themselves to learn what comprises a realistic face, and in so doing, devise an encoding (a latent space) used as a starting point for subsequent syntheses. The optimization usually consumes millions of authentic face images, whose features are implicitly incorporated into the discriminator and thus into the resulting model. The model is then sampled to generate images of the faces of ``nonexistent'' people, starting with a latent space vector. There two important research questions at hand, which we address in this paper: 

\begin{quote}
{\it Q1. Does identity information from the real subjects, as embodied in training data, leak into the generated (“fake”) samples?}
\end{quote}

\begin{quote}
{\it Q2. If answer to Q1 is affirmative, might this be a threat to privacy?}
\end{quote}

Figure \ref{fig:overview} shows the flow of this information as it passes through the model training process from real image to fake image. So, if by chance, one's face image data exists in the training data, there is an opportunity for their data to inadvertently leak. These newly generated faces should then reproduce training identities in a way that skews the performance of a face recognition system. That is, the probability of being falsely matched to a face that "does not exist" is higher than when one's face is not included in the training data set. This paper is the first known to us to conduct experimental assessment of this problem in StyleGAN \cite{karras2019style}, and we demonstrate that {\bf identity information leakage does exist} for some face matchers while for others it does not. This variation in information leakage across matchers also demands careful consideration in the context of face recognition system design.


This result stimulates additional intriguing research questions: can or should we sample the latent space where realistic face representations reside in some specific way to avoid such information leakage when generating face images of ``nonexistent'' subjects? Is there any specific arrangement of authentic faces in this latent space? If so, can we learn the general characteristics and geometry of this realistic manifold? We hope the results presented in this paper will stimulate a scientific discourse about the role of generative models in biometrics and their privacy-related attributes.

We offer databases of real and synthetic faces used in this work to other researchers to facilitate replication of our experiments and investigation of new related problems.

\section{Related Work}

\subsection{Generative Models}

\begin{figure}[t]
    \centering
    \includegraphics[width=.48\textwidth]{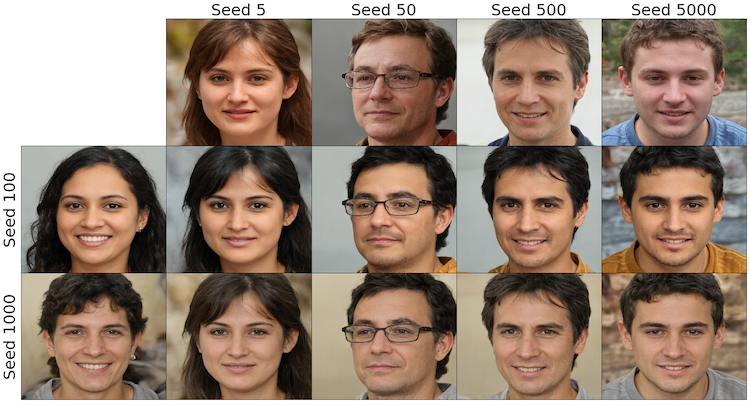}
    \caption{Style Mixing with StyleGAN2 (Truncation $\psi=0.5$)}
    \label{fig:mixing}
\end{figure}

A primary motivation for the use of generative models is their ability to produce high-quality data. To this end, NVIDIA's most recent contribution to the world of generative models is called StyleGAN2 \cite{karras2019analyzing}, which improves upon their original StyleGAN work \cite{karras2019style}. Before StyleGAN, NVIDIA released ProGAN \cite{karras2017progressive}, one of the first GANs to produce high-quality images. Given a large enough training data set, all three of these models can be trained to reproduce representative images of nearly any object. However, one of the most popular, successful applications has been a given model's ability to generate high-resolution face images that are informally considered very realistic. Built atop the ProGAN paradigm and trained on the FFHQ data set \cite{karras2019style}, both StyleGAN and StyleGAN2 learned to recreate these synthetic faces from the ground up using separable resolution-based ``styles,'' hence the name. Coarse styles, also referred to as features, control pose and general face shape; medium styles control more detailed hair and facial features; fine features control color scheme for eyes, hair, skin, and even background. Due to the designed separability of these learned features, hybrid faces can be synthesized through latent vector manipulation, while still exhibiting real image characteristics, with some degree of useful control, as seen in Figure \ref{fig:mixing}. In this figure, each ``inner'' cell can be seen as a combination of the coarse features of the column and the medium-to-fine features of the row. 

\subsection{Information Leakage}

GAN-based leakage and model inversion attacks have proven to be rather effective in terms of extracting sensitive data from generative models, both intentionally and unintentionally. Hitaj \etal \cite{hitaj2017deep} suggest that GANs can act like composite artists generating sketches of an eyewitness’s memory of an individual. Though this ``sketch'' (a synthetic sample) may not encompass all features of the individual, relevant and distinctive details about this person could be exposed. This is further explained by Fredrikson \etal \cite{fredrikson2015model} who focus on adversarial attacks. Given access to a generative model's architecture and weights, they show that one can leverage facial recognition confidence scores to reconstruct a person's face image. Goodfellow \etal \cite{goodfellow2014explaining} assert that a GAN (given infinite time and resources) would generate samples originally found in the training data, \ie there is a possibility (albeit small) for an unprovoked 100\% information leakage for a given sample from the training set.

\subsection{Remedies for Information Leakage}

Since sensitivity regarding personal data leakage into machine learning models has been so well documented, several solutions have been developed to prevent such information leakage \cite{xu2015privacy, beaulieu2019privacy, xin2020private}. At the heart of many such efforts is the notion of differential privacy, which acts as a privacy guarantee for sensitive data in the context of machine learning. For GANs, differential privacy can ensure synthetic data production with a limited risk of private information exposure. Xu \etal \cite{xu2019ganobfuscator} offer GANobfuscator as a solution to the problem of data leakage in GANs. GANobfuscator adopts differential privacy in two ways: Gaussian noise injection and gradient-pruning; the latter method also boosts model stability during the training process. Once the model has been trained on the noise-infused data, ``new" data can be generated by GANobfuscator and released into the public with mathematically bounded concern for privacy attacks, such as the already mentioned reverse engineering of training data. Similarly, Chen \etal \cite{chen2018differentially} offer 
GANs based on auto-encoders and variational auto-encoders designed to protect privacy while retaining high data utility.

\begin{table*}[t]
    \centering\small
    \caption{Face Matcher Details}
    \begin{tabular}{c|c|c|c|c}
    \hline
    \hline
    Model Name & Architecture & Training Data (\# of images) & Feature Vector Length & Comparison Score\\
    \hline
    face\_recognition \cite{fr} & ResNet-34 & Custom ($\sim$3M) & 128 & Euclidean Distance\\
    FaceNet \cite{schroff2015facenet} & GoogLeNet-24 & VGGFace2 \cite{Cao18} ($\sim$3.3M) & 512 &  Euclidean Distance\\
    ArcFace \cite{deng2019arcface} & ResNet-100 & MS-Celeb-1M \cite{guo2016msceleb} ($\sim$10M) & 512 & Cosine Distance\\
    SphereFace \cite{liu2017sphereface} & ResNet-64 & CASIA-Webface \cite{yi2014learning} ($\sim$0.45M) & 512 & Cosine Distance\\
    MegaMatcher 11.2 \cite{neuro} & Proprietary & Proprietary & Proprietary & Proprietary \\
    \hline
    \end{tabular}
    \label{tab:face_matcher_details}
\end{table*}

\subsection{Fake Image Detection}

Substantial efforts have gone into ``Deepfake" detection, which asks the question of whether an image has been digitally created or manipulated in some way \cite{hulzebosch2020detecting}, \cite{matern2019exploiting}, \cite{neves2020ganprintr}, \cite{tolosana2020deepfakes}. A famous example of Deepfakes follows a ``This {\it X} Does Not Exist" pattern, wherein {\it X} could be cats, apartment units, cars, or faces. For each of these examples, a separate GAN has been trained to generate fake images of the designated object. As mentioned earlier, GANs are incredibly good at doing this convincingly. In response to fake image generation is fake image detection. Further, researchers have studied the effects of GAN-generated fake face images on face recognition systems \cite{korshunov2018deepfakes}, claiming that several state of the art face matchers are vulnerable to Deepfake videos.

However, this paper aims not to determine whether a face image is fake. Quite contrarily, we aim to prove that within these generated fake images are remnants of the real images ingested during the GAN training process.

\section{Data Sets}

We use three data sets in experiments presented in this paper. The first data set is the Flickr-Faces-HQ data set (FFHQ), which includes 70,000 high-resolution ($1024 \times 1024$) images of human faces varying in age, ethnicity, image background, and facial clothing/apparel (glasses, hats, etc). The entirety of the FFHQ data was used by NVIDIA labs to train the StyleGAN2 model, as well as the original StyleGAN. Each of 70,000 images in FFHQ represents a different individual's face.

The second data set we used is a collection of 70,000 StyleGAN2-generated fake face images developed in this work and offered along with the paper. We will call this data set ``SG2''. Along with several pre-trained models, NVIDIA provides scripts to generate faces via latent vector manipulation. In our case, we used latent vectors of length 512, corresponding to seed values from 0 through 69999. StyleGAN also adjusts latent vectors using the truncation trick to balance image quality and image variety. Truncation values of 0.0 generate high-quality images near the center, or mean of the latent space, whereas truncation values of 1.0 generate high-variation images from peripheral areas of poorer representation in the latent space. 
For our synthetic image data set (SG2), we used a truncation value of $\psi=0.5$ to generate variety-quality balanced face images. The provided generator script can be found in NVIDIA's repository: \url{https://github.com/NVlabs/stylegan2}.


The third data set, used solely to assess the face recognition accuracy of various face matchers used in this work, is an assortment of 12,004 high quality face images collected by the authors. The data set features 333 subjects with between 4 and 78 front-facing photos per subject, captured at a resolution of $1200\times 1600$ under good lighting. This data, further referred to as \emph{UNI}, was used to gauge the relative performance of our face matchers. This set is subject-disjoint with both the FFHQ and SG2 databases.


\section{Methodology \& Experiments}

In this paper, ``identity leakage" refers to the flow of identity-salient facial features from real training data images (in FFHQ) to fake synthetic images (in SG2). We hypothesize that these identity-salient features are readily detectable by face matchers, whose job is to judge the similarity or dissimilarity of two face images. 

For a given pair of images, a face matcher can quantitatively compare the two images via extracted face feature vectors. This comparison between face images is commonly expressed as a distance, where larger distances correspond to dissimilar images and smaller distances correspond to similar images. To prove our hypothesis, we want to show that the distance between a real image and a generated image is, on average, smaller than the distance between two real images. If this {\it is} the case, then identity ``information" (identity-salient features) from the training data must be spilling into the generated data, as evidenced by the smaller distances between extracted feature vectors.

Since not all face matchers use dissimilarity-based distance scores, we simply refer to comparisons of images as ``comparison scores". Comparison scores between pairs of real images are referred to as ``R-R" (Real-Real) while comparison scores between one real image and one generated image are referred to as ``R-G" (Real-Generated). Since these scores inherently rely on the extracted face feature vectors, identity leakage might be observable for some face matchers but not others.


\begin{figure*}[t]
    \includegraphics[width=\textwidth]{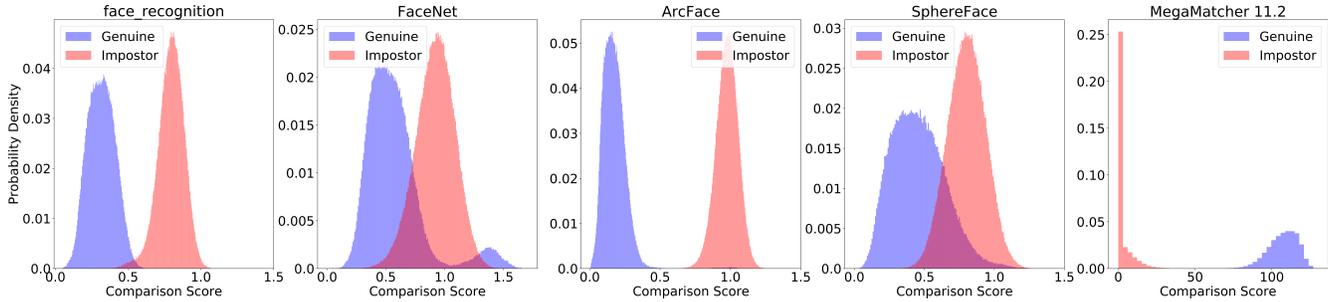}
    \caption{Genuine and impostor score distributions for each matcher on {\it UNI}. From left to right: face\_recognition, FaceNet, ArcFace, SphereFace, MegaMatcher 11.2.}
    \label{fig:uni_histograms}
\end{figure*}

\subsection{Face Matcher Details}

We used five face matchers to verify our hypothesis about identity information leakage. Four of the five matchers were open source, CNN-based, and implemented in Python. The fifth technique (MegaMatcher 11.2) was a commercial face matcher from Neurotechnology implemented in C++. Information regarding model architecture and training data can be found in Table \ref{tab:face_matcher_details}. Note that MegaMatcher 11.2 uses similarity-based scores, \ie lower scores correspond to dissimilar face images while higher scores correspond to similar face images. All other selected face matchers used distance, or dissimilarity scoring (lower scores indicate stronger similarity).

\begin{figure*}[b]
    \includegraphics[width=\textwidth]{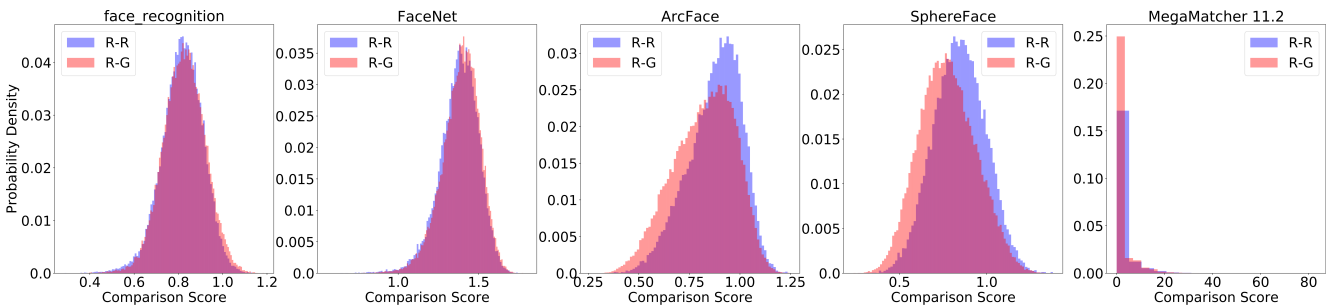}
    \caption{R-R and R-G score distributions for each matcher. From left to right: face\_recognition, FaceNet, ArcFace, SphereFace, MegaMatcher 11.2.}
    \label{fig:realgen_histograms}
\end{figure*}

\subsection{Preliminary Performance Analysis on UNI}

We conducted an initial set of experiments on our \emph{UNI} data set to determine the relative performance of the selected face matchers, to -- potentially -- link the strength of the identity leakage with the accuracy of a given face matcher. The first step of our pipeline was to extract face feature vectors from all 12,004 face images. In order to enable a fair comparison, we removed from the experiment all images that failed the face detection step in any of the five matchers. This yielded a usable image subset of 11,792 images.

Since the \emph{UNI} data set contains multiple images per subject, matching experiments yielding both genuine and impostor comparison scores were conducted. The genuine scores numbered 290,122, while the impostor scores numbered 442,224. Genuine scores were calculated using all images available in the data set. On the other hand, impostor scores were calculated using a subset of the data. This subset was comprised of four images per subject since all subjects in the \emph{UNI} data set had at least four images. A given subject's ``first" two images (indices 0 and 1) were compared against the two ``latter" images (indices 2 and 3) of all other subjects. So, in total there were 333 subjects $\times$ 2 ``first" images $\times$ 332 other subjects $\times$ 2 ``latter" images = 442,224 impostor comparison scores per matcher. This methodology was developed to reduce computation time as well as balance the number of images per subject in the data set. Comparison score results as well as the failure to enroll rate (FTE) for each matcher can be seen in Table \ref{tab:uni_scores}. Impostor and genuine score distributions are shown in Figure \ref{fig:uni_histograms}.

\begin{table}[h]
    \centering
    \caption{\emph{UNI} Comparison Results}
    \begin{tabular}{c|c|c|c}
    \hline
    \hline
    Model Name & FTE (\%) & AUC & $d'$ \\ [0.5ex]
    \hline
    face\_recognition & 9.99e-4 & 0.9991 & 5.1930 \\
    FaceNet & 2.49e-4 & 0.8945 & 1.6048 \\
    ArcFace & 2.49e-4 & 0.9978 & 9.5513 \\
    SphereFace & 2.49e-4 & 0.9261 & 2.1124 \\
    MegaMatcher 11.2 & 1.76e-2 & 0.9999 & 13.6908 \\
    \hline
    \end{tabular}
    \label{tab:uni_scores}
\end{table}


To quantify the similarity and dissimilarity between scoring distributions, we use the decidability measure $d'$, which cumulates sample mean and sample variances of comparison scores into a scalar factor:

\begin{equation}
d' = \frac{|\smean{x}_g-\smean{x}_i|}{\sqrt{\frac{1}{2}\big(\svar{x}_g+\svar{x}_i\big)}}
\label{eqn:dprime}
\end{equation}

\noindent
where $\smean{x}$ and $\svar{x}$ denote sample mean and sample variance of a comparison score $x$, respectively. 
The value of $d'$ estimates the degree by which the genuine distribution of $x_g$ and impostor distribution of $x_i$ overlap (the higher $d'$ is, the lower is the overlap). For uncorrelated random variables, $d'=\sqrt{2}\zeta$ where $\zeta$ is a standardized difference between the expected values of two random variables, often used in hypothesis testing.

Based on $d'$ (as seen in Table \ref{tab:uni_scores}), the relative ranking of the five matchers is: MegaMatcher 11.2 $>$ ArcFace $>$ face\_recognition $>$ SphereFace $>$ FaceNet.

\subsection{Impostor Distribution Analysis between FFHQ \& SG2}

Two impostor distributions of comparison scores were produced for each face matcher. The first shows the distribution of ``Real-Real'' (R-R) comparison scores; the second shows the distribution of ``Real-Generated'' (R-G) comparison scores. 

The R-R partition was created by pairing real images from the FFHQ data set together. Although FFHQ originally included 70,000 real images, 12,906 images failed to enroll in one or more of the matchers and were removed from the set of usable ``real" images. The remaining 57,094 images were split in half and matched together for a total of 28,547 pairs.

The R-G partition was created by pairing each of the usable 57,094 ``real" images with an image from the usable ``generated" SG2 data set, i.e. the set of synthetic images that properly enrolled in all five matchers. Note that 628 SG2 images failed to enroll to at least one matcher. 

Figure \ref{fig:realgen_histograms} illustrates the R-R and R-G comparison score distributions for each face matcher. Additionally, FTE for each matcher for each data set, $d'$ and the two-sample Kolmogorov-Smirnov test statistic are reported in Table \ref{tab:table3}. Similar to $d'$, the K-S test quantifies the distance between two distributions; the larger the test statistic is, the greater is the distance.

\begin{table*}[t]
    \centering\small
    \caption{``Real-real faces'' (R-R) and ``Real- generated faces'' (R-G) comparison results, along with failure-to-enroll rates for each face matcher, separately for real samples FTE$_{R}$ and generated samples FTE$_{G}$.}
    \begin{tabular}{c|c|c|c|c|c|c}
    \hline
    \hline
    Model Name & FTE$_{R}$ & FTE$_{G}$ & $d'$ & K-S Test Statistic & $p$-value & Difference in Means (R-R$\rightarrow$R-G) \\
    \hline
    face\_recognition & 3.62e-3 & 0.0 & 0.0855 & 0.0332 & $<$0.001 & $+$0.008082\\
    FaceNet & 1.71e-4 & 0.0 & 0.1055 & 0.0496 & $<$0.001 & $+$0.01266288 \\
    ArcFace & 1.71e-4 & 0.0 & 0.3801 & 0.1562 & $<$0.001 & $-$0.05384\\
    SphereFace & 1.71e-4 & 0.0 & 0.4404 & 0.1895 & $<$0.001 & $-$0.06824\\
    MegaMatcher 11.2 & 1.84e-1 & 8.97e-3 & 0.0116 & 0.0026 & 0.9978 & $-$0.04813\\
    \hline
    \end{tabular}
    \label{tab:table3}
\end{table*}

The first immediate and interesting observation (from Table \ref{tab:table3}) is that FTE values are lower for synthetic images that from real images. This may be because face detectability in its output is ``baked into'' the network, by limiting its training inputs to images with detectable faces. Future work might investigate how truncation value ($\psi$) affects face detection and recognition for StyleGAN2-generated face images.

The second interesting observation is that we see a {\bf noticeable shift between the R-R and R-G impostor distributions towards genuine scores} for ArcFace and SphereFace matchers in Fig. \ref{fig:realgen_histograms}. This type of shift suggests that the generated face images do in fact contain identity-salient features from the training data, \ie identity leakage is evident. We elaborate more on this observation in the next section. Note that although the observed differences are statistically significant ($p < 0.001$ at the significance level $alpha=0.01$), the difference in means suggests that only ArcFace and SphereFace have leftward shifts from R-R to R-G. On the other hand, the face\_recognition package and FaceNet model show statistically significant shifts to the right for the R-G distribution, which actually suggests the opposite of identity leakage. This varying behavior between matchers is curious and will be further discussed the next section as well.

\section{Results \& Discussion}

\subsection{Answering Q1: Does identity information from the real subjects leak into the generated samples?}

Before conducting any experiments, we hypothesized that results from {\it all} five selected face matchers would support the existence of identity leakage. However, we quickly found that this was not the case after generating and comparing the R-R and R-G impostor distributions with face\_recognition, one of the quicker off-the-shelf face matchers. 

We then adjusted our hypothesis, thinking that maybe only the strongest performing face matchers would support our leakage claim, hence the need for ranking the face matchers on the \emph{UNI} data set. There is a precedent for this assumption in the iris recognition domain, where the author found that the most accurate matchers are to the highest extent sensitive to ``biometric template aging'' \cite{10.1007/978-3-662-44485-6_20}. Based on the relative ranking of the face matchers (by the $d'$ metric), we suspected that the MegaMatcher 11.2 SDK and the ArcFace face matchers would yield leakage-evident results. 

However, as seen in Table \ref{tab:table3}, identity leakage is supported only in ArcFace and SphereFace, the second and fifth face matcher in terms of accuracy, respectively. Figure \ref{fig:realgen_histograms} offers visual support to the same conclusions. In the two leftmost plots, we see that the face\_recognition and FaceNet matchers do not show signs of identity leakage, as is the case with MegaMatcher 11.2 in the rightmost plot. There is strong overlap between the R-R and R-G comparison score distributions for all three of these matchers, suggesting minimal or no information leakage. Contrarily, for ArcFace and SphereFace, the R-G score distributions are shifted to the left of the R-R scores. Since ArcFace and SphereFace comparison scores are distance-based, a shift to the left (smaller distances) {\it does} support identity leakage as such a shift hints that generated face images are, on average, ``closer" to real face images.

\begin{figure*}[t]
    \centering
    \begin{subfigure}[h]{.36\textwidth}
        \centering
        \includegraphics[width=.92\linewidth]{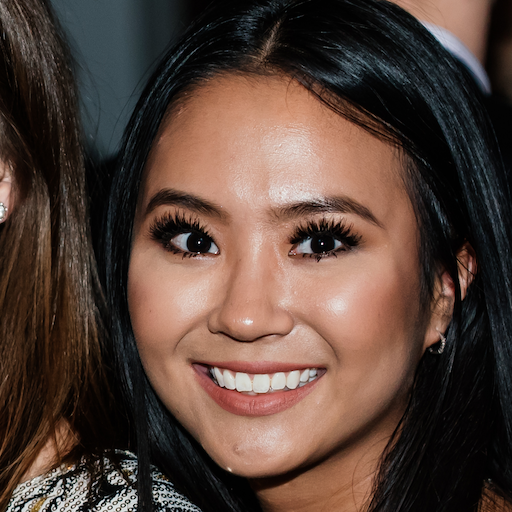}
        \caption{FFHQ/51243.png}
        \label{fig:pic1}
    \end{subfigure}
    \begin{subfigure}[h]{.36\textwidth}
        \centering
        \includegraphics[width=.92\linewidth]{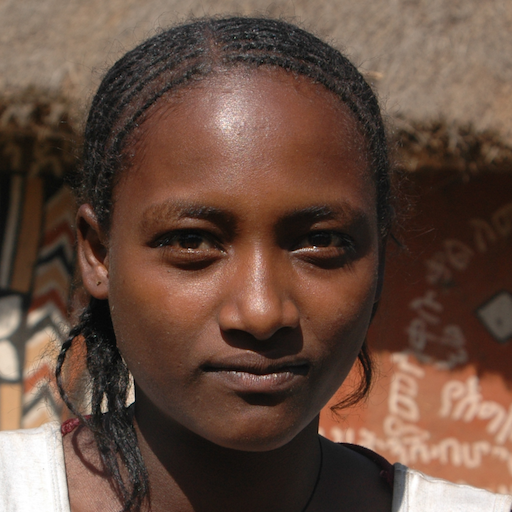}
        \caption{FFHQ/14788.png}
        \label{fig:pic2}
    \end{subfigure}
    
    \vspace{2mm}
    
    \begin{subfigure}[h]{.36\textwidth}
        \centering
        \includegraphics[width=.95\linewidth]{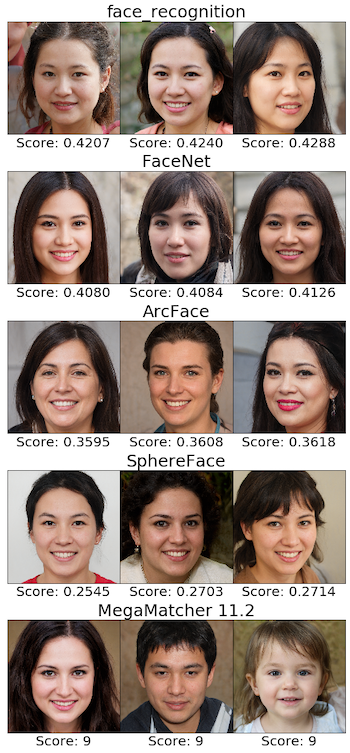}
        \caption{The closest samples to (a)}
        \label{fig:grid1}
    \end{subfigure}
    \begin{subfigure}[h]{.36\textwidth}
        \centering
        \includegraphics[width=.95\linewidth]{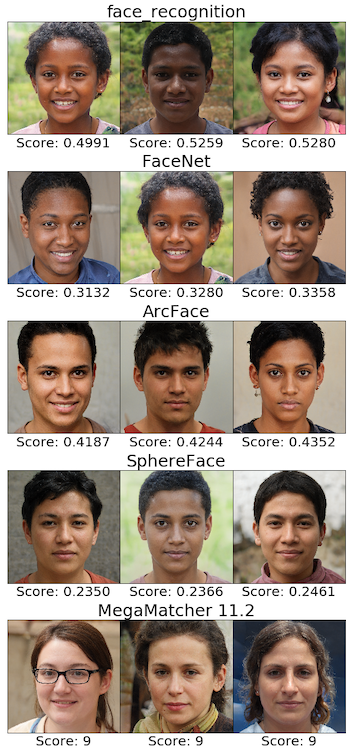}
        \caption{The closest samples to (b)}
        \label{fig:grid2}
    \end{subfigure}
    \caption{Three closest fake face images (SG2) to real face images from FFHQ shown in (a) and (b) according to each face matcher: face\_recognition, FaceNet, ArcFace, SphereFace, MegaMatcher 11.2. Row images are sorted left-to-right in increasing order of distance to the corresponding base image.}
    \label{fig:face_grids}
\end{figure*}

The question then becomes: {\bf why does identity leakage only appear for certain face matchers?} What information is leaking that leads to difference in impostor comparison scores? To better understand these inquiries from a qualitative standpoint, we constructed ``closest image" grids in Figure \ref{fig:face_grids}. The objective of these grids is to investigate how different face matchers judge ``close" face images, and in so doing, gain insight into why only some matchers think fake face images look more like the face images from specific real identities. The base images (\ref{fig:pic1}, \ref{fig:pic2}) are two real face images from the FFHQ data set. The grids below each base image (\ref{fig:grid1}, \ref{fig:grid2}) show rows of images for each face matcher. Each row showcases the three closest synthetic face images from the SG2 data to the corresponding base image at the top, with the closest image on the left and farthest on the right. For all matchers but MegaMatcher 11.2, comparison score is a distance measure (Euclidean distance for face\_recognition and FaceNet, and cosine distance for ArcFace and SphereFace). For these matchers, lower scores correspond to closer or more similar face images. For MegaMatcher, higher comparison scores mean more similar face images. 

Upon observing Figure \ref{fig:face_grids}, we note a few points of interest regarding the face matchers' ability to match fake and real face images. Comparison scores for MegaMatcher 11.2 (bottom row) do not offer much in terms of visual correspondence between face images. Inaccurate scores span race, gender, and age. This suggests that perhaps MegaMatcher 11.2 relies on more complex, less obvious facial features to compare, match, and verify faces. For reference, to achieve a false acceptance rate of 0.0001\%, the matching threshold is 72, which is 8x the score (9) of the two examples in \ref{fig:face_grids}. Since the software is proprietary and comparison results are rather stagnant for non-match image pairs, there is not much we can discuss in terms of leakage. 

Between the remaining four models, there seem to be two groups: matchers that display evidence of identity leakage (ArcFace and SphereFace), and matchers that do not (face\_recognition and FaceNet). The first group, simply based on nomenclature, seems to revolve around nonlinear geometry. In \cite{deng2019arcface}, Deng \etal explicitly compare ArcFace and SphereFace. Though the former improves upon the latter, they retain a few identifying similarities. Both apply angular margin penalties to boost performance for intra-class appearance variation, such as extreme pose and age difference. ArcFace uses additive angular margin while SphereFace uses multiplicative angular margin. Further, both matchers compare face feature vectors via trigonometric functions and share a ResNet model architecture. Perhaps these similarities may play a role in explaining why these models seem to support identity leakage while other models do not. Further work to this end includes using GradCAM visualization \cite{selvaraju2017grad} to quantitatively discuss face matcher behavior regarding real and fake images.

\vspace{3mm}

Contrary to ArcFace and SphereFace, the face\_recognition and FaceNet matchers seem to favor linearity. One of the main focuses of Google's contribution with FaceNet was Euclidean-based triplet loss. Similarly, the face\_recognition matcher measures similarity between face feature vectors in Euclidean distance. We have yet to explore this rigorously, but perhaps linearity versus non-linearity in face matcher geometry has an effect on identity leakage.

\clearpage
Circling back to our initial question, we have experimentally shown that {\bf the answer to Q1 is affirmative: the identity information \underline{can} leak into generated data, although the strength of this phenomenon is uneven across the matchers}.

\subsection{Answering Q2: Might the observed data leakage a threat to privacy?}

\begin{figure}[h]
    \includegraphics[width=.47\textwidth]{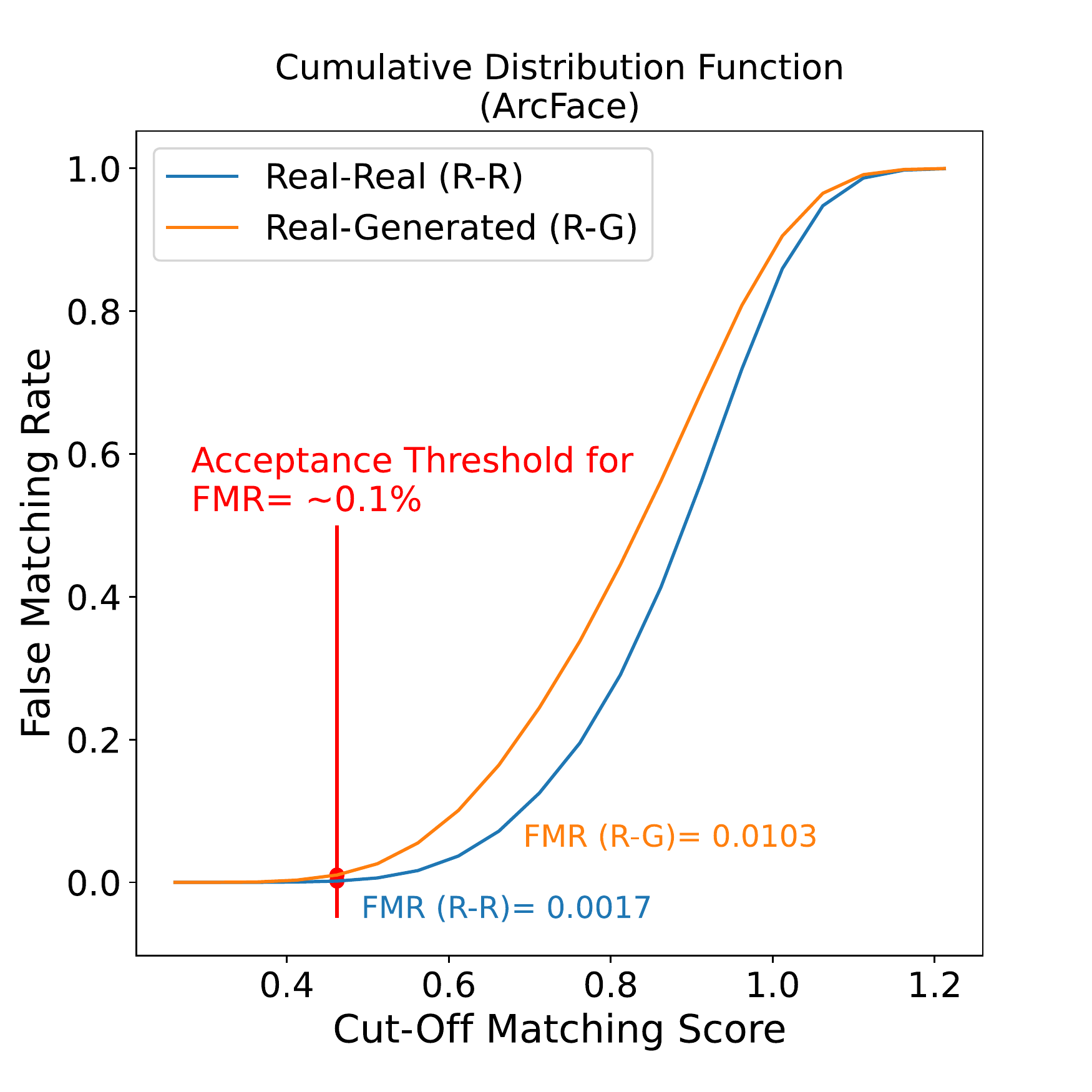}
    \caption{Illustration of how false match rate probability increases when our face image is included in the training set of StyleGAN2 with ArcFace-based matching}
    \label{fig:cdf}
\end{figure}

Figure \ref{fig:cdf} shows the false matching rate as a function of the acceptance threshold for comparison scores between real and unrelated-real (R-R), and real vs potentially-related-generated (R-G) face images when using the ArcFace face matcher (the method prone to the identity leakage to the highest extent in our experiments). We can note a clear shift in false match probability when ``unexisting'' faces are being matched against real subjects. To better understand this plot, we can discuss a social media-inspired example. Recently, Facebook shut down thousands of Facebook and Instagram accounts that used ``AI-generated images'' as profile photos. If Facebook had chosen to use a matcher behaving as shown in Fig. \ref{fig:cdf} to hunt down these fake profile photos (as generated by StyleGAN2), then real users whose face images appeared in the training data (yellow curve) would be {\bf 6.6 times more likely to be falsely matched} when compared to someone who did not appear in the training data (blue curve). The baseline false positive rate comes from a NIST standard of 0.001 (1 in 1000) for face recognition. Thus, {\bf the answer to Q2 is also affirmative: there does exist a practical threat to privacy}, and its seriousness may be encountered even for matchers presenting solid identification performance for authentic face images.

\subsection{Inconclusiveness}
The fact that our findings are split between proving and disproving identity leakage emphasizes the importance of careful and thoughtful design in face matching systems. Why some facial features trickle through one face recognition model while not through others is still unknown. Accordingly, practical use cases of face recognition need to account for the fact that generative models can be used to exploit these models without an obvious adversarial attack. Further work, including several iterations of retraining of StyleGAN2, will hopefully shed more light on the causes and further effects of identity leakage in generative models.

\section{Conclusions and Future Directions}

Identity leakage in generative models requires careful attention, if only for the potential impact on privacy. It has been shown in previous works that sensitive data can be pried out of models via adversarial attacks, but in this paper we have also shown that in some cases this data can simply spill out of the training corpus into synthetic samples without any provocation. We also demonstrated that identity leakage depends on the face matcher technology in use, which may be a potentially impactful result for privacy protection-aware network design used in future models. As mentioned earlier, future endeavors include the usage of network visualization tools, such as GradCAM or activation atlases, for quantitative studies of matcher behavior in the context of identity information leakage. Illustrating the differences between face feature extraction for identity-leaking and identity-preserving models would certainly help us understand what features are used for comparing faces. Are these the same type of facial features that can be leaked? Additionally, between the combinations of training data sets, model architectures, and loss functions, there are many, many more flavors of face matcher to test for identity leakage. Among those are CosFace, DeepFace, VGGFace, and SphereFace+. In a similar vein, we intend to also explore other face synthesizer models, such as StarGAN, CycleGAN, and GauGAN. Finally, a validation of proposed differential privacy-based GANs on an even larger set of synthetic face images would test the capabilities of available solutions and offer feedback to better prevent unwanted identity leakage.

{\small
\bibliographystyle{ieee}
\bibliography{main}

\begin{thebibliography}{10}\itemsep=-1pt

\bibitem{fr}
{face\_recognition: The world's simplest facial recognition API for Python and
  the command line}.
\newblock https://github.com/ageitgey/face\_recognition.

\bibitem{neuro}
Neurotechnology: Fingerprint, face, eye iris, voice and palm print
  identification, speaker and object recognition software.
\newblock https://www.neurotechnology.com/.

\bibitem{beaulieu2019privacy}
B.~K. Beaulieu-Jones, Z.~S. Wu, C.~Williams, R.~Lee, S.~P. Bhavnani, J.~B.
  Byrd, and C.~S. Greene.
\newblock Privacy-preserving generative deep neural networks support clinical
  data sharing.
\newblock {\em Circulation: Cardiovascular Quality and Outcomes},
  12(7):e005122, 2019.

\bibitem{Cao18}
Q.~Cao, L.~Shen, W.~Xie, O.~M. Parkhi, and A.~Zisserman.
\newblock {VGGFace2: A dataset for recognising faces across pose and age}.
\newblock In {\em International Conference on Automatic Face and Gesture
  Recognition}, 2018.

\bibitem{chen2018differentially}
Q.~Chen, C.~Xiang, M.~Xue, B.~Li, N.~Borisov, D.~Kaarfar, and H.~Zhu.
\newblock Differentially private data generative models.
\newblock {\em arXiv preprint arXiv:1812.02274}, 2018.

\bibitem{10.1007/978-3-662-44485-6_20}
A.~Czajka.
\newblock Influence of iris template aging on recognition reliability.
\newblock In M.~Fern{\'a}ndez-Chimeno, P.~L. Fernandes, S.~Alvarez, D.~Stacey,
  J.~Sol{\'e}-Casals, A.~Fred, and H.~Gamboa, editors, {\em Biomedical
  Engineering Systems and Technologies}, pages 284--299, Berlin, Heidelberg,
  2014. Springer Berlin Heidelberg.

\bibitem{deng2019arcface}
J.~Deng, J.~Guo, N.~Xue, and S.~Zafeiriou.
\newblock Arcface: Additive angular margin loss for deep face recognition.
\newblock In {\em Proceedings of the IEEE Conference on Computer Vision and
  Pattern Recognition}, pages 4690--4699, 2019.

\bibitem{fredrikson2015model}
M.~Fredrikson, S.~Jha, and T.~Ristenpart.
\newblock Model inversion attacks that exploit confidence information and basic
  countermeasures.
\newblock In {\em Proceedings of the 22nd ACM SIGSAC Conference on Computer and
  Communications Security}, pages 1322--1333, 2015.

\bibitem{goodfellow2014generative}
I.~Goodfellow, J.~Pouget-Abadie, M.~Mirza, B.~Xu, D.~Warde-Farley, S.~Ozair,
  A.~Courville, and Y.~Bengio.
\newblock Generative adversarial nets.
\newblock In {\em Advances in neural information processing systems}, pages
  2672--2680, 2014.

\bibitem{goodfellow2014explaining}
I.~J. Goodfellow, J.~Shlens, and C.~Szegedy.
\newblock Explaining and harnessing adversarial examples.
\newblock {\em arXiv preprint arXiv:1412.6572}, 2014.

\bibitem{gulrajani2017improved}
I.~Gulrajani, F.~Ahmed, M.~Arjovsky, V.~Dumoulin, and A.~C. Courville.
\newblock {Improved training of Wasserstein GANs}.
\newblock In {\em Advances in neural information processing systems}, pages
  5767--5777, 2017.

\bibitem{guo2016msceleb}
Y.~Guo, L.~Zhang, Y.~Hu, X.~He, and J.~Gao.
\newblock M{S}-{C}eleb-1{M}: A dataset and benchmark for large scale face
  recognition.
\newblock In {\em European Conference on Computer Vision}. Springer, 2016.

\bibitem{hitaj2017deep}
B.~Hitaj, G.~Ateniese, and F.~Perez-Cruz.
\newblock Deep models under the gan: information leakage from collaborative
  deep learning.
\newblock In {\em Proceedings of the 2017 ACM SIGSAC Conference on Computer and
  Communications Security}, pages 603--618, 2017.

\bibitem{hulzebosch2020detecting}
N.~Hulzebosch, S.~Ibrahimi, and M.~Worring.
\newblock Detecting cnn-generated facial images in real-world scenarios.
\newblock In {\em Proceedings of the IEEE/CVF Conference on Computer Vision and
  Pattern Recognition Workshops}, pages 642--643, 2020.

\bibitem{karras2017progressive}
T.~Karras, T.~Aila, S.~Laine, and J.~Lehtinen.
\newblock {Progressive growing of GANs for improved quality, stability, and
  variation}.
\newblock {\em arXiv preprint arXiv:1710.10196}, 2017.

\bibitem{karras2019style}
T.~Karras, S.~Laine, and T.~Aila.
\newblock A style-based generator architecture for generative adversarial
  networks.
\newblock In {\em Proceedings of the IEEE Conference on Computer Vision and
  Pattern Recognition}, pages 4401--4410, 2019.

\bibitem{karras2019analyzing}
T.~Karras, S.~Laine, M.~Aittala, J.~Hellsten, J.~Lehtinen, and T.~Aila.
\newblock {Analyzing and improving the image quality of StyleGAN}.
\newblock {\em arXiv preprint arXiv:1912.04958}, 2019.

\bibitem{korshunov2018deepfakes}
P.~Korshunov and S.~Marcel.
\newblock Deepfakes: a new threat to face recognition? assessment and
  detection.
\newblock {\em arXiv preprint arXiv:1812.08685}, 2018.

\bibitem{liu2017sphereface}
W.~Liu, Y.~Wen, Z.~Yu, M.~Li, B.~Raj, and L.~Song.
\newblock Sphereface: Deep hypersphere embedding for face recognition.
\newblock In {\em Proceedings of the IEEE conference on computer vision and
  pattern recognition}, pages 212--220, 2017.

\bibitem{matern2019exploiting}
F.~Matern, C.~Riess, and M.~Stamminger.
\newblock Exploiting visual artifacts to expose deepfakes and face
  manipulations.
\newblock In {\em 2019 IEEE Winter Applications of Computer Vision Workshops
  (WACVW)}, pages 83--92. IEEE, 2019.

\bibitem{neves2020ganprintr}
J.~C. Neves, R.~Tolosana, R.~Vera-Rodriguez, V.~Lopes, H.~P. Proen{\c{c}}a, and
  J.~Fierrez.
\newblock Ganprintr: Improved fakes and evaluation of the state of the art in
  face manipulation detection.
\newblock {\em IEEE Journal of Selected Topics in Signal Processing}, 2020.

\bibitem{radford2015unsupervised}
A.~Radford, L.~Metz, and S.~Chintala.
\newblock Unsupervised representation learning with deep convolutional
  generative adversarial networks.
\newblock {\em arXiv preprint arXiv:1511.06434}, 2015.

\bibitem{schroff2015facenet}
F.~Schroff, D.~Kalenichenko, and J.~Philbin.
\newblock Facenet: A unified embedding for face recognition and clustering.
\newblock In {\em Proceedings of the IEEE conference on computer vision and
  pattern recognition}, pages 815--823, 2015.

\bibitem{selvaraju2017grad}
R.~R. Selvaraju, M.~Cogswell, A.~Das, R.~Vedantam, D.~Parikh, and D.~Batra.
\newblock Grad-cam: Visual explanations from deep networks via gradient-based
  localization.
\newblock In {\em Proceedings of the IEEE international conference on computer
  vision}, pages 618--626, 2017.

\bibitem{shen2019fake}
C.~Shen, M.~Kasra, W.~Pan, G.~A. Bassett, Y.~Malloch, and J.~F. O’Brien.
\newblock Fake images: The effects of source, intermediary, and digital media
  literacy on contextual assessment of image credibility online.
\newblock {\em new media \& society}, 21(2):438--463, 2019.

\bibitem{tolosana2020deepfakes}
R.~Tolosana, R.~Vera-Rodriguez, J.~Fierrez, A.~Morales, and J.~Ortega-Garcia.
\newblock Deepfakes and beyond: A survey of face manipulation and fake
  detection.
\newblock {\em arXiv preprint arXiv:2001.00179}, 2020.

\bibitem{xin2020private}
B.~Xin, W.~Yang, Y.~Geng, S.~Chen, S.~Wang, and L.~Huang.
\newblock Private fl-gan: Differential privacy synthetic data generation based
  on federated learning.
\newblock In {\em ICASSP 2020-2020 IEEE International Conference on Acoustics,
  Speech and Signal Processing (ICASSP)}, pages 2927--2931. IEEE, 2020.

\bibitem{xu2019ganobfuscator}
C.~Xu, J.~Ren, D.~Zhang, Y.~Zhang, Z.~Qin, and K.~Ren.
\newblock Ganobfuscator: Mitigating information leakage under gan via
  differential privacy.
\newblock {\em IEEE Transactions on Information Forensics and Security},
  14(9):2358--2371, 2019.

\bibitem{xu2015privacy}
K.~Xu, H.~Yue, L.~Guo, Y.~Guo, and Y.~Fang.
\newblock Privacy-preserving machine learning algorithms for big data systems.
\newblock In {\em 2015 IEEE 35th international conference on distributed
  computing systems}, pages 318--327. IEEE, 2015.

\bibitem{yi2014learning}
D.~Yi, Z.~Lei, S.~Liao, and S.~Z. Li.
\newblock Learning face representation from scratch.
\newblock {\em arXiv preprint arXiv:1411.7923}, 2014.

\end{thebibliography}
}

\end{document}